\documentclass[a4paper]{styles/svproc}
\usepackage{url}

\usepackage{xcolor}
\usepackage{amssymb}
\usepackage{graphicx}
\usepackage{mathtools}
\usepackage{amsmath}
\usepackage{gensymb}
\usepackage{float}
\usepackage{multirow}
\usepackage{makecell}
\usepackage{mathtools}
\usepackage{hyperref}
\hypersetup{
    colorlinks=true,
    linkcolor=blue,
    filecolor=magenta,      
    urlcolor=blue,
    pdftitle={Overleaf Example},
    pdfpagemode=FullScreen,
    }

\usepackage{tabularx}
    \newcolumntype{L}{>{\raggedright\arraybackslash}X}
\usepackage[utf8]{inputenc}
\usepackage[english]{babel}

\usepackage{amsthm}

\usepackage{algorithm}
\usepackage{algpseudocode}

\usepackage{subcaption}

\newcommand{\no}{\noindent}

\begin{document}
\mainmatter              

\title{AdVENTR: Autonomous Robot Navigation in Complex Outdoor Environments}
%
\titlerunning{Outdoor Navigation}  
%
\author{Kasun Weerakoon,  Adarsh Jagan Sathyamoorthy, Mohamed Elnoor \and
Dinesh Manocha }
\authorrunning{Weerakoon et al.} 
%
\tocauthor{Kasun Weerakoon, Adarsh Jagan Sathyamoorthy, Mohamed Elnoor, Dinesh Manocha}
\institute{University of Maryland, College Park, MD, USA,
}

\maketitle              

\begin{abstract}
We present a novel system, AdVENTR for autonomous robot navigation in unstructured outdoor environments that consist of uneven and vegetated terrains. Our approach is general and can enable both wheeled and legged robots to handle outdoor terrain complexity including unevenness, surface properties like poor traction, granularity, obstacle stiffness, etc. We use data from sensors including RGB cameras, 3D Lidar, IMU, robot odometry, and pose information with efficient learning-based perception and planning algorithms that can execute on edge computing hardware. Our system uses a scene-aware switching method to perceive the environment for navigation at any time instant and dynamically switches between multiple perception algorithms. We test our system in a variety of sloped, rocky, muddy, and densely vegetated terrains and demonstrate its performance on Husky and Spot robots.



\end{abstract}

\section{Introduction}
\begin{figure}[t]
    \centering
    \includegraphics[width=\columnwidth,height=2.4cm]{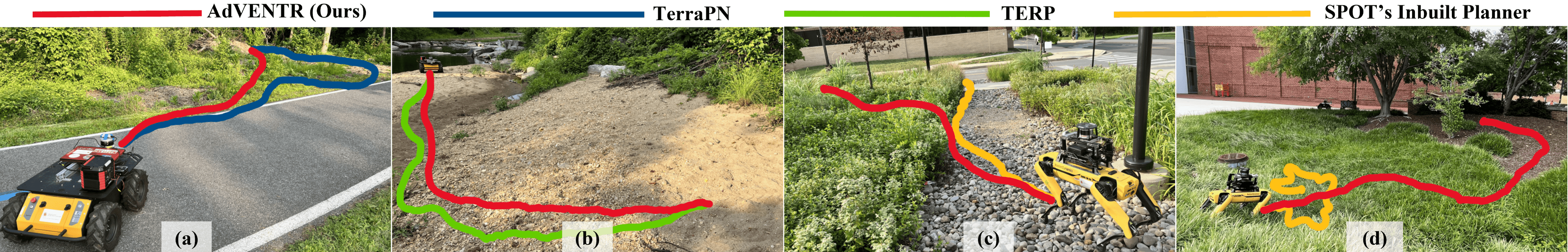}
    \caption{\small{Navigation trajectories generated by our approach and methods in comparison on diversely challenging outdoor terrains : (a) Asphalt and pliable vegetation terrains; (b) Uneven and sandy terrains; (c) Granular and partially pliable vegetation terrains; (d) Uneven and pliable vegetation; for wheeled (a Husky in scenarios (a) and (b)) and legged robot (a Spot in scenarios (c) and (d)) platforms. }}
    \label{fig:traj_figs}
    \vspace{-16pt}
\end{figure}



In recent years, robots have been used for outdoor applications such as seeding, harvesting, crop health measurement in agriculture \cite{harvesting-robots}, search and rescue \cite{niroui2019deep}, jobsite monitoring in construction \cite{afsari2021fundamentals}, delivery, etc. Additionally, many new possibilities have emerged due to the recent advances in legged robots that are capable of traversing challenging outdoor terrains.  However, unstructured outdoor environments presents a variety of challenges for autonomous navigation such as uneven terrains, surface undulations, granularity (e.g. sand, mud, snow, etc), vegetation with a variety of stiffness properties (e.g tall grass, trees), buildings, etc.~\cite{4598867,sathyamoorthy2023vern}. In addition, the resulting systems need to take into account:  1. the robot's dynamics constraints;  2. limited onboard computational capabilities for efficient perception and navigation. 


Our system is designed for both wheeled and quadruped robots. A wheeled robot may be able to traverse slopes and handle rocky, slippery, and granular surfaces to a certain extent. However, such robot's would fail in steep or vertical slopes (e.g. stairs), extremely granular sand or mud where their wheels could slip, or in dense grass even if it is safe to plough through them~\cite{4598867}. On the other hand, legged robots possess higher degrees of freedom which allows them to successfully traverse all the aforementioned terrains and more. Therefore, any method for perception and navigation must account for such differences in robot capabilities by differentiating traversable and non-traversable paths for each robot.

Secondly, robots are typically equipped with limited onboard computation resources, that need to process the sensor data in realtime for perception and navigation computations.  Existing model-based navigation approaches \cite{fox1997dynamic,orca,overbye1} have a relatively low computational overhead. However, they address only collision avoidance with respect to rigid or stiff obstacles in flat, smooth terrains and may not work weel in uneven, rough terrains. Recently, machine learning-based approaches for perception and navigation have been used in more challenging scenarios~\cite{chen2018deep,faust2018prm,wellhausen2021rough,weerakoon2023terrain}. However, developing a unified, end-to-end learning-based pipeline to address all the outdoor challenges is non-trivial for real-world scenarios because of the challenges of fusing highly diverse multi-modal data, high computational overhead~\cite{tang2022perception,weerakoon2023graspe} and lack of sufficient training data. Based on these current challenges, we frame our approach's problem statement as follows:


\textbf{Problem Statement:} \textit{To develop a versatile, modular, computationally-light outdoor navigation system capable of navigating wheeled and legged robots in uneven, undulated, granular, and densely vegetated terrains.}


\section{Proposed Method}
\vspace{-5pt}

We propose, \textbf{AdVENTR} an \textbf{Ad}aptive \textbf{V}egetation, \textbf{E}levation, a\textbf{N}d \textbf{T}errain-aware \textbf{R}obot navigation strategy. In our system architecture, we decompose autonomous outdoor robot navigation into robot perception and planning. Our perception module is used for interpreting and understanding the environmental components such as terrain elevation, surface condition (poor traction, granularity, etc), and pliability/stiffness of the vegetation around the robot using raw sensory data. However, identifying such environmental conditions using a unified or single perception approach is non-trivial due to the complexity of the input multi-modal data, the scale of the perception model, and  high computational requirements of the learning methods. Hence, we use three separate perception sub-modules. These modules compute cost maps that are used to quantify terrain traversability. Since the performance of each perception sub-module varies from one environment to another, we propose a scene-aware switching mechanism to alternate between perception sub-modules to improve the accuracy and computational efficiency. Moreover, we propose a planner that can generate navigation actions by incorporating the cost maps provided by the perception modules. Our system uses sensor modalities such as 3D point clouds, RGB images, Inertial Measurement Units (IMU), the robot's pose, velocity, and joint positions and forces (for a legged robot). The overall system architecture is shown in Fig. \ref{fig:sys_arch}.

\subsection{Sensory Inputs}
We utilize an RGB camera, a 3D LiDAR, Odometry, a 6-DOF IMU (for wheeled robots), and joint feedback and motor current(for legged robots) sensors to perform scene understanding and navigability estimation during perception. We acquire the sensor data streams: RGB image ($I_t^{RGB}$), 3D point cloud ($P_t^{lidar}$), robot's position w.r.t the odometry frame ($x_t^O, y_t^O$), current linear and angular velocities ($v_t, \omega_t$), its orientation (i.e, roll ($\phi_t$), pitch ($\theta_t$), yaw ($\psi_t$) angles), IMU data vector ($V_t^{IMU}$), and leg joint feedback vector ($J_{t}$) for a legged robot at a given time $t$ to conduct several scene understanding and perception tasks summarized in the following sub-sections.

\subsection{Scene-aware Switching} \label{sec:scene-understand}
The scene-aware switching quantifies the terrain challenges in the robot's surroundings using different pre-processed sensory observations to choose an appropriate combination of perception modules. In particular, we use three metrics to quantify the unevenness, surface properties, and the existence of vegetation to decide the suitable combination for perception modules for a scenario. At any time, our system can choose a maximum of two perception modules to execute by prioritizing them based on the severity of the terrain challenges, i.e. to optimize the computations. We use the below three measures to obtain conditions for switching between the perception modules:

\begin{figure}[t]
    \centering
    \includegraphics[width=0.8\columnwidth,height=4cm]{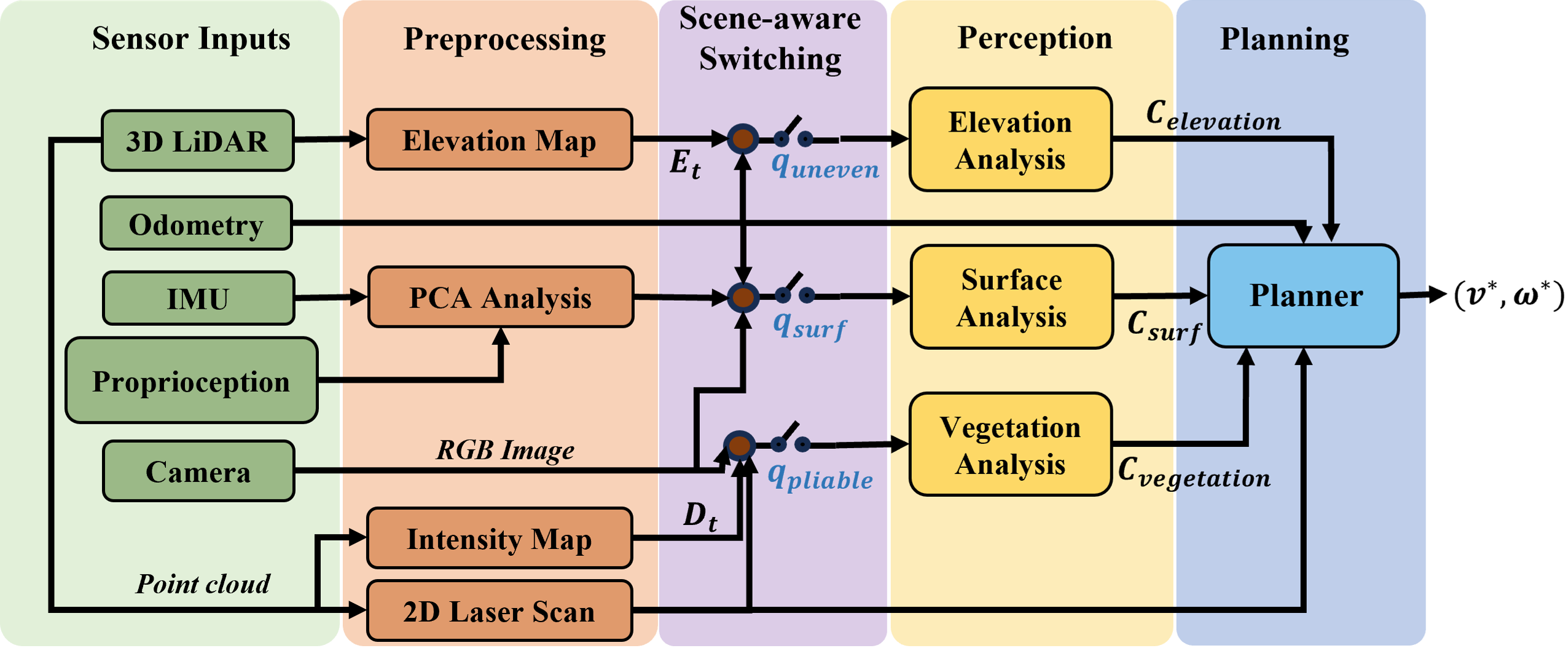}
    \caption{\small{The overall system architecture of our outdoor navigation framework displaying how our scene-aware switching module (section \ref{sec:scene-understand}) switches between multiple perception methods depending on the predominant challenge in the outdoor environment. Our outdoor navigation planner integrates the outputs of one or a combination of two perception methods to compute an optimal, collision-free velocity. Additionally, our framework supports both wheeled and legged robots.}}
    \label{fig:sys_arch}
\end{figure}

\begin{figure}[t]
    \centering
    \includegraphics[width=\columnwidth]{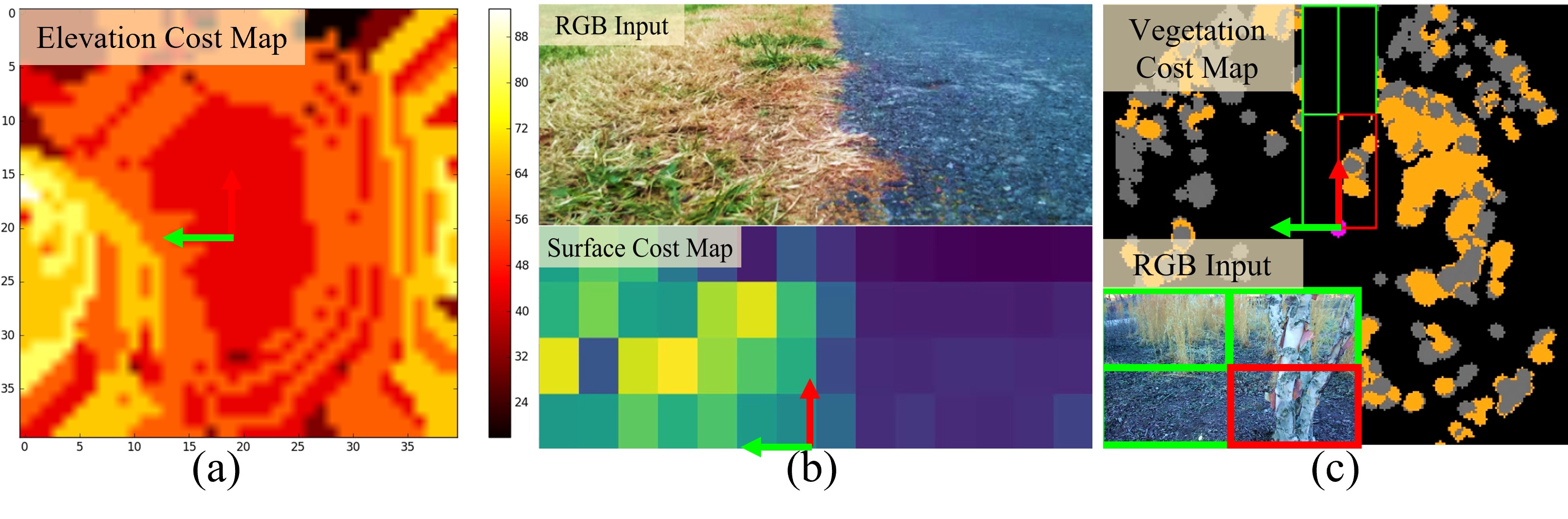}
    \caption{\small{Perception Cost Maps Generated from LiDAR and camera that represents terrain elevations, roughness, and solidity of the obstacles. The robot-centric elevation cost map is generated from the 3D LiDAR point cloud to estimate the elevation changes of the robot's vicinity. The surface cost map in (b) is generated from TerraPN \cite{sathyamoorthy2022terrapn} (for wheeled robots) by correlating the instability experienced by the robot from IMU and odometry drift with the RGB image view of the robot. The vegetation cost map generated by VERN \cite{sathyamoorthy2023vern} incorporates a vision-based vegetation classifier and a LiDAR-based occupancy map as shown in (c).}}
    \label{fig:cost_maps}
    \vspace{-5pt}
\end{figure}

\textbf{Terrain Unevenness ($q_{uneven}$)}: We use a robot-centric elevation map $E_t \in [0,100]^{N \times N}$ derived from 3D point clouds, and calculate the gradient of the elevation. If the maximum gradient is greater than a threshold, we consider elevation to be a predominant terrain challenge. 

\textbf{Surface Properties ($q_{surf}$)}: We apply principal component analysis to the IMU data (for a wheeled robot), and joint position, forces (for a legged robot) and observe the variances of the data along the first two dimensions $\sigma_{PC_1}, \sigma_{PC_2}$ (see Fig. \ref{fig:scene_und_fig}a). If $\sqrt{\sigma_{PC_1}^2+ \sigma_{PC_2}^2}$ is greater than a threshold, we consider surface properties to be a predominant terrain challenge. 

\textbf{Vegetation ($q_{pliable}$)}: To estimate the presence of vegetation in the environment, we use a map ($D_t \in [0,100]^{N \times N}$) that represents the reflecting intensity (directly proportional to the density/solidity of obstacles) of the LiDAR rays in the robot's vicinity. Intermediate intensities ($\sim 50$) in $D_t$ are used to estimate the presence of pliable vegetation such as tall grass. Additionally, we check for the presence of vertical gradients in $E_t$ to indicate the obstacles. If both conditions are satisfied, we consider handling vegetation to be the predominant challenge for the planning module.

If several or all of these switching conditions are satisfied, we utilize combinations of perception methods based on the following priority: Vegetation $>$ Unevenness $>$ Surface properties. Finally, our method detects sensor streams corresponding to a wheeled (e.g. wheel odometry) and a legged robot (e.g. joint positions) and automatically uses only the methods applicable to its dynamics. 
\vspace{-10pt}

\subsection{Outdoor Perception} \label{sec:perception}
Based on the predominant terrain challenges at a given time, our system chooses to execute a pair of the perception methods below to handle them. We leverage four perception modules introduced in our previous work: TERP \cite{weerakoon2022terp}, TerraPN \cite{sathyamoorthy2022terrapn}, ProNAV \cite{elnoor2023pronav}, and VERN \cite{sathyamoorthy2023vern}. 

\subsubsection{Terrain Elevation Perception with TERP: }
To perceive the geometric properties of outdoor terrains, i.e., their slope, unevenness, etc., we use TERP \cite{weerakoon2022terp}, a method that uses lidar point clouds, the robot's pose, and goal to compute an elevation-aware cost map of the environment. TERP trains an end-to-end DRL model that uses an elevation map $E$ (obtained by processing point clouds) to compute an attention mask $A$ that can be applied over the input elevation map $E$. The resulting masked cost map $(E \times A)$ identifies and highlights the reduced-stability regions where the robot could have unstable poses (pitch and roll angles) or even topple (see Fig. \ref{fig:cost_maps}(a)).

\subsubsection{Terrain Surface Perception with TerraPN and ProNAV :} Outdoor navigation challenges arise from a terrain's surface properties such as traction, bumpiness, granularity, etc can cause undesirable effects such as high vibrations, wheel/leg slip, entrapment in granular surfaces such as sand, mud, etc. To perceive such surface properties, we propose two methods: 1. TerraPN \cite{sathyamoorthy2022terrapn} for wheeled robots, and 2. ProNav \cite{elnoor2023pronav} for legged robots. TerraPN \cite{sathyamoorthy2022terrapn} learns the surface properties (traction, bumpiness, deformability, etc.) of complex outdoor terrains directly from robot-terrain interactions through self-supervised learning. It uses RGB images of terrain surfaces and the robot's velocities as inputs, and the IMU vibrations and wheel odometry errors experienced by the robot as labels for self-supervision. TerraPN outputs a \textit{surface} cost map relative to the RGB image that differentiates smooth, high-traction surfaces (low navigation costs) from bumpy, slippery, granular surfaces (high navigation costs) as shown in Fig. \ref{fig:cost_maps}(b). 


\subsubsection{Vegetation Perception with VERN: } To identify dense vegetation, differentiate pliable/traversable objects (such as tall grass) and non-pliable/hard obstacles such as trees, we use VERN \cite{sathyamoorthy2023vern}. VERN trains a vision-based classifier on quadrants divided from RGB camera images to classify vegetation in the scene as tall grass, bushes, or trees. The classification results along with its confidence, and vegetation height (extracted from lidar point clouds) are used to calculate a vegetation-aware cost map (see Fig. \ref{fig:cost_maps}(c)).







\subsection{Planner}
\vspace{-5pt}
Our framework uses a novel, versatile planner capable of interfacing with the perception methods described in section \ref{sec:perception}. We extend the Dynamic Window Approach (DWA) \cite{fox1997dynamic} with a novel objective function to compute the maximizing linear and angular velocities $(v^*, \omega^*)$ for the robot's feasible velocity set.
\vspace{-5pt}
\begin{gather}
    G(v, \omega) =  \alpha.\text{Head}.(1 - \textbf{Surface}) + \beta.\text{Obs} + \gamma.\text{Vel} + \delta.\textbf{Elevation} + \theta.\textbf{Veg}.
    \label{eqn:objective-func}
\end{gather}
\no In this case, Head() measures the progress towards the robot's goal, Obs() is the distance to the closest obstacle when executing a certain $(v, \omega)$,  Vel() measures the forward velocity of the robot and encourages higher velocities, Surface() denotes the surface cost from obtained from TerraPN, Elevation() represents the elevation cost from TERP, and Veg() is the vegetation cost from VERN. 
The highlighted terms are added newly as part of our formulation. They are used depending on our scene understanding module which activates a perception module based on the detected navigation challenge in the environment.

\section{Results and Evaluations}
\begin{figure}[t]
    \centering
    \includegraphics[width=\columnwidth,height=3.5cm]{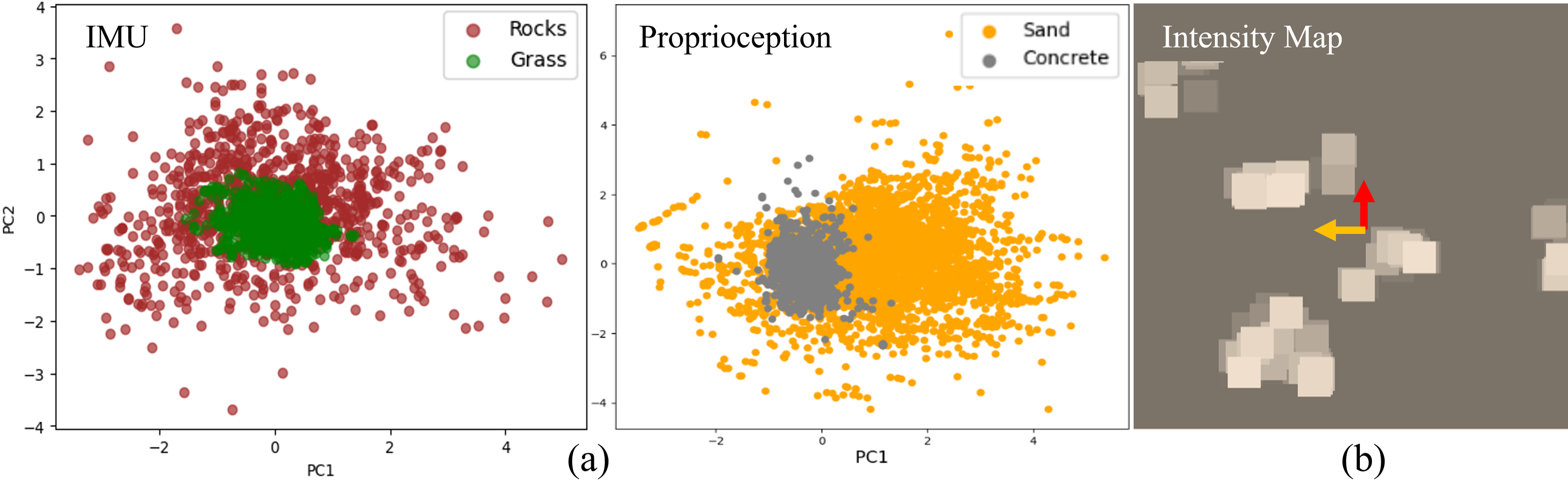}
    \caption{\small{Sensor data analysis for scene-aware switching. (a) PCA analysis on IMU data for wheeled, and joint position and force data for legged robots respectively. Granular terrains such as rocks (brown) and sand (orange) lead to high variances ($\sigma_{PC_1}, \sigma_{PC_2}$) in the first two principal components. (b) LiDAR point cloud intensity map used to identify the existence of soft/pliable objects (i.e., white squares).}}
    \label{fig:scene_und_fig}
\end{figure}

\vspace{-5pt}
We evaluate our system's navigation performance in several challenging real-world environments using a Clearpath Husky robot (wheeled), and a Boston Dynamics Spot robot (legged). The robots are equipped with a Velodyne VLP16 lidar, a Realsense L515 camera, and a computer with an Nvidia RTX 20xx series GPU. The preprocessing results (shown in Fig. \ref{fig:scene_und_fig}) are used to perform scene-aware switching. In our evaluations, we consider four test scenarios where only two out of the three aforementioned perception challenges (i.e., unevenness, surface condition, and vegetation) are dominant. Then, our system chooses a suitable combination of perception methods (e.g. TerraPN + VERN, TERP + TerraPN, TERP + VERN, ProNav + VERN, etc), which we compare against the standalone perception approaches. We summarize our preliminary qualitative and quantitative navigation performance in Fig. \ref{fig:traj_figs} and Table \ref{tab:comparisons} respectively. 

\begin{table}[ht]

\begin{center}       
\begin{tabular}{|c|c|c|c|c|c|}
\hline
\rule[-1ex]{0pt}{2.5ex}  Scenario & Method &  Success Rate  & Mean & Instability  & Elevation \\
\rule[-1ex]{0pt}{2.5ex}   &  &  (\%) & Velocity & Cost & Gradient \\
\hline
\rule[-1ex]{0pt}{2.5ex}  Scenario 1 & TerraPN \cite{sathyamoorthy2022terrapn} & 60 & 0.402 & 1.129 & 0.378  \\
\rule[-1ex]{0pt}{2.5ex}   & AdVENTR (ours) &  \textbf{90} & 0.427 & 1.142 & 0.355 \\
\hline
\rule[-1ex]{0pt}{2.5ex}  Scenario 2 & TERP \cite{weerakoon2022terp}  & 80 & 0.531 & 1.367 & 1.326 \\
\rule[-1ex]{0pt}{2.5ex}   & AdVENTR (ours) & \textbf{100} & 0.425 & 1.174 & 1.297 \\
\hline
\rule[-1ex]{0pt}{2.5ex}  Scenario 3 & Spot's Inbuilt  & 0  &  0.438 & 16.2 & 1.024\\
\rule[-1ex]{0pt}{2.5ex}   & AdVENTR (ours) & \textbf{80} & 0.553 & 7.23 & 0.872\\
\hline
\rule[-1ex]{0pt}{2.5ex}  Scenario 4 & Spot's Inbuilt  & 10  &  0.412 & 11.63 & 2.375\\
\rule[-1ex]{0pt}{2.5ex}   & AdVENTR (ours) & \textbf{90} & 0.483 & 5.3 & 1.944\\
\hline
\rule[-1ex]{0pt}{2.5ex}  Scenario 5 & Spot's Inbuilt  & 20  &  0.438 & 12.38 & 2.438\\
\rule[-1ex]{0pt}{2.5ex}   & AdVENTR (ours) & \textbf{90} & 0.503 & 6.2 & 2.067\\
\hline
\rule[-1ex]{0pt}{2.5ex}  Scenario 6 & VERN \cite{sathyamoorthy2023vern}  & 80  &  0.536 & 9.76 & 0.998\\
\rule[-1ex]{0pt}{2.5ex}   & AdVENTR (ours) & \textbf{90} & 0.544 & 5.7 & 0.765\\
\hline
\end{tabular}
\end{center}
\caption{\small{Navigation performance comparisons in terms of various metrics. Instability cost is measured using the cumulative vibrations experienced by the robot's IMU while navigating to a goal. Elevation cost is the summation of the elevation gradients experienced by the robot along a trajectory. }}
\label{tab:comparisons}
\end{table}

\section{Experimental Insights and Discussion}
\begin{figure}[t]
    \centering
    \includegraphics[width=\columnwidth,height=6cm]{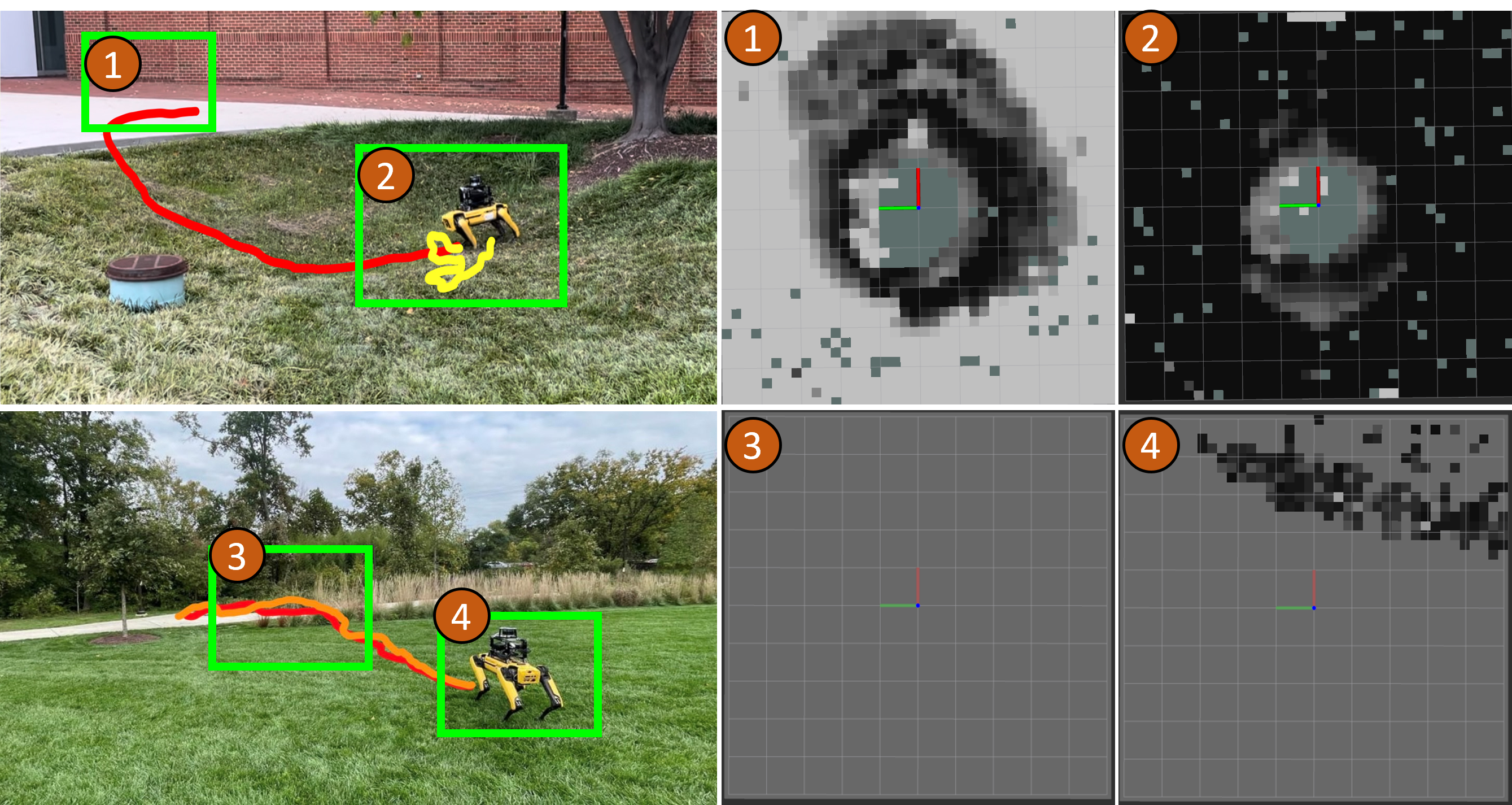}
    \caption{\small{Navigation trajectories in Scenario 5 \textbf{[Top]} and Scenario 6 \textbf{[Bottom]} and sample visualization of elevation and intensity maps that capture unevenness and vegetation for scene-aware switching along a trajectory. In scenario 5, elevation perception from TERP is used to navigate since the high unevenness is reflected in the elevation map (i.e., \textbf{[Top Middle]}). However, at the later part of the trajectory, the robot navigates to a flat region where elevation-aware navigation using TERP is unnecessary. Similarly, in scenario 6 \textbf{[Bottom]}, the existence of vegetation is reflected in the intensity map \textbf{[Bottom Right]} which encourages to use of VERN during the later part of the trajectory.}}
    \label{fig:nav_comp2}
\end{figure}

\textbf{Scenario 1} in Fig. \ref{fig:traj_figs}(a) consists of asphalt and pliable vegetation where a wheeled robot can navigate through. However, surface-aware perception from TerraPN considers vegetation regions as obstacles to avoid which results in wandering and longer trajectories. In contrast, our method incorporates both TerraPN and VERN to identify pliable vegetation to generate significantly shorter trajectories at a higher success rate. 

\textbf{Scenario 2} in Fig. \ref{fig:traj_figs}(b) is an uneven sandy/granular terrain where the robots encounter wheel slips and irregular traction. In such scenarios, TERP can only recognize the elevation properties whereas TerraPN only identifies the granularity of the surface. Hence, TERP generates relatively higher velocities to navigate along the elevated regions which results in heavy wheel slips and odometer errors. Alternatively, our method combines TerraPN with TERP where the navigation velocity adaptively changes to reduce the slipping.

\textbf{Scenario 3} in Fig. \ref{fig:traj_figs}(c) includes a rocky surface, pliable vegetation, and trees. We observe that our legged robot's inbuilt controller considers the traversable vegetation as obstacles and attempts to avoid it during all the trials (i.e., zero success rate). In contrast, our method incorporates ProNav to change to a stable gait on rocks and VERN to identify pliable vegetation to reach the goal with a significantly lower instability cost.

\textbf{Scenario 4} in Fig. \ref{fig:traj_figs}(d) is an uneven/sloped region that includes grass and mulch. Spot's inbuilt planner leads to highly inconsistent ground estimations which eventually results in catastrophic robot failures (crashes). However, our method incorporates TERP to navigate in the path with the least elevation gain and VERN to identify pliable grass regions and walk through them, leading to a significantly higher success rate.

\textbf{Scenario 5} in Fig. \ref{fig:nav_comp2}\textbf{[Top]} is an extension of scenario 4. However, the robot also encounters flat concrete ground after facing slopes with grass. AdVENTR initially uses a combination of TERP and VERN similar to scenario 4 and then switches to ProNav due since in the absence of elevation and vegetation, the surface properties could become the predominant challenge for navigation. However, Spot's inbuilt planner performs equivalent to scenario 4 and crashes in most trials.  

\textbf{Scenario 6} in Fig. \ref{fig:nav_comp2}\textbf{[Bottom]} has the robot operating on flat lawns, then moving through tall pliable grass, then on to concrete pavements, and finally on to concrete pavement again. Our method accurately detects that at subsequent intervals of time, surface properties and vegetation are the challenges that need to be handled. Therefore, it initially uses only ProNav, and when the robot approaches the vegetation, AdVENTR uses both ProNav and VERN. We compare the performance of our method with VERN. Qualitatively, the trajectories, and quantitatively, the success rates and elevation gradients are comparable. However, our method leads to a lower mean velocity and instability cost as ProNav switches to the most stable, slower gait when required.     


\section{Conclusions, Limitations and Future Work}
In this work, we presented AdVENTR, a method that unifies multiple perception sub-modules to deal with various outdoor navigation challenges such as elevation changes, surface properties, and dense vegetation. We demonstrated this unified method on real-world wheeled and legged robots in various challenging outdoor terrains. Our method has a few limitations. AdVENTR's performance can be affected by motion blur, occlusions, and lighting changes since its perception sub-modules (TerraPN and VERN) use RGB images. Moreover, low-hanging objects (e.g., branches and leaves) can be misrepresented in the terrain elevation map. To achieve better generalization in novel environments, lighting and weather conditions in the future, we plan to use Vision Language Models (VLMs) to understand the scene better and switch perception modules. Additionally, the perception modules themselves can be improved/augments using VLMs. 


\section{Acknowledgements}

This work was supported in part by ARO Grants W911NF2110026, W911NF2310046,  W911NF2310352  and Army Cooperative Agreement W911NF2120076.


%

%
%
%
\bibliographystyle{splncs04}
\bibliography{references}

\begin{thebibliography}{10}
\providecommand{\url}[1]{\texttt{#1}}
\providecommand{\urlprefix}{URL }
\providecommand{\doi}[1]{https://doi.org/#1}

\bibitem{afsari2021fundamentals}
Afsari, K., Halder, S., Ensafi, M., DeVito, S., Serdakowski, J.: Fundamentals
  and prospects of four-legged robot application in construction progress
  monitoring. EPiC Series in Built Environment  \textbf{2},  274--283 (2021)

\bibitem{orca}
van~den Berg, J., Guy, S., Lin, M., Manocha, D.: Reciprocal n-Body Collision
  Avoidance, vol.~70, pp. 3--19 (04 2011). \doi{10.1007/978-3-642-19457-31}

\bibitem{chen2018deep}
Chen, X., Ghadirzadeh, A., Folkesson, J., Bj{\"o}rkman, M., Jensfelt, P.: Deep
  reinforcement learning to acquire navigation skills for wheel-legged robots
  in complex environments. In: 2018 IEEE/RSJ international conference on
  intelligent robots and systems (IROS). pp. 3110--3116. IEEE (2018)

\bibitem{elnoor2023pronav}
Elnoor, M., Sathyamoorthy, A.J., Weerakoon, K., Manocha, D.: Pronav:
  Proprioceptive traversability estimation for autonomous legged robot
  navigation in outdoor environments. arXiv preprint arXiv:2307.09754  (2023)

\bibitem{faust2018prm}
Faust, A., Oslund, K., Ramirez, O., Francis, A., Tapia, L., Fiser, M.,
  Davidson, J.: Prm-rl: Long-range robotic navigation tasks by combining
  reinforcement learning and sampling-based planning. In: 2018 IEEE
  international conference on robotics and automation (ICRA). pp. 5113--5120.
  IEEE (2018)

\bibitem{fox1997dynamic}
Fox, D., Burgard, W., Thrun, S.: The dynamic window approach to collision
  avoidance. IEEE Robotics \& Automation Magazine  \textbf{4}(1),  23--33
  (1997)

\bibitem{harvesting-robots}
Kootstra, G., Wang, X., Blok, P.M., Hemming, J., Van~Henten, E.: Selective
  harvesting robotics: current research, trends, and future directions. Current
  Robotics Reports  \textbf{2},  95--104 (2021)

\bibitem{4598867}
Low, C.B., Wang, D.: Gps-based tracking control for a car-like wheeled mobile
  robot with skidding and slipping. IEEE/ASME Transactions on Mechatronics
  \textbf{13}(4),  480--484 (2008). \doi{10.1109/TMECH.2008.2000827}

\bibitem{niroui2019deep}
Niroui, F., Zhang, K., Kashino, Z., Nejat, G.: Deep reinforcement learning
  robot for search and rescue applications: Exploration in unknown cluttered
  environments. IEEE Robotics and Automation Letters  \textbf{4}(2),  610--617
  (2019)

\bibitem{overbye1}
Overbye, T., Saripalli, S.: Path optimization for ground vehicles in off-road
  terrain. In: 2021 IEEE International Conference on Robotics and Automation
  (ICRA). pp. 7708--7714 (2021). \doi{10.1109/ICRA48506.2021.9561291}

\bibitem{sathyamoorthy2022terrapn}
Sathyamoorthy, A.J., Weerakoon, K., Guan, T., Liang, J., Manocha, D.: Terrapn:
  Unstructured terrain navigation using online self-supervised learning. In:
  2022 IEEE/RSJ International Conference on Intelligent Robots and Systems
  (IROS). pp. 7197--7204. IEEE (2022)

\bibitem{sathyamoorthy2023vern}
Sathyamoorthy, A.J., Weerakoon, K., Guan, T., Russell, M., Conover, D., Pusey,
  J., Manocha, D.: Vern: Vegetation-aware robot navigation in dense
  unstructured outdoor environments. arXiv preprint arXiv:2303.14502  (2023)

\bibitem{tang2022perception}
Tang, Y., Zhao, C., Wang, J., Zhang, C., Sun, Q., Zheng, W.X., Du, W., Qian,
  F., Kurths, J.: Perception and navigation in autonomous systems in the era of
  learning: A survey. IEEE Transactions on Neural Networks and Learning Systems
   (2022)

\bibitem{weerakoon2023graspe}
Weerakoon, K., Sathyamoorthy, A.J., Liang, J., Guan, T., Patel, U., Manocha,
  D.: Graspe: Graph based multimodal fusion for robot navigation in outdoor
  environments. IEEE Robotics and Automation Letters  (2023)

\bibitem{weerakoon2023terrain}
Weerakoon, K., Sathyamoorthy, A.J., Manocha, D.: Terrain-aware autonomous robot
  navigation in outdoor environments. In: Open Architecture/Open Business Model
  Net-Centric Systems and Defense Transformation 2023. vol. 12544, pp. 78--85.
  SPIE (2023)

\bibitem{weerakoon2022terp}
Weerakoon, K., Sathyamoorthy, A.J., Patel, U., Manocha, D.: Terp: Reliable
  planning in uneven outdoor environments using deep reinforcement learning.
  In: 2022 International Conference on Robotics and Automation (ICRA). pp.
  9447--9453. IEEE (2022)

\bibitem{wellhausen2021rough}
Wellhausen, L., Hutter, M.: Rough terrain navigation for legged robots using
  reachability planning and template learning. In: 2021 IEEE/RSJ International
  Conference on Intelligent Robots and Systems (IROS). pp. 6914--6921. IEEE
  (2021)

\end{thebibliography}
\end{document}